\pdfoutput=1

\documentclass[11pt]{article}

\usepackage[final]{acl2023}

\usepackage{times}
\usepackage{latexsym}

\usepackage[T1]{fontenc}

\usepackage[utf8]{inputenc}
\usepackage[section]{placeins}
\usepackage{microtype}

\usepackage{inconsolata}

\usepackage{array}
\usepackage{graphicx}
\usepackage{multirow}

\makeatletter
\newcommand*{\textoverline}[1]{$\overline{\hbox{#1}}\m@th$}
\makeatother

\title{Iry\textoverline{o}NLP\thanks{The team name \textit{iry\textoverline{o}} comes from the japanese for medical or healthcare.} at MEDIQA-CORR 2024:\\Tackling the Medical Error Detection \& Correction Task\\On the Shoulders of Medical Agents}

\author{Jean-Philippe Corbeil \\
  Microsoft Health \& Life Sciences \\
  \texttt{jcorbeil@microsoft.com}
}

\begin{document}

\maketitle

\begin{abstract}
In natural language processing applied to the clinical domain, utilizing large language models has emerged as a promising avenue for error detection and correction on clinical notes, a knowledge-intensive task for which annotated data is scarce. This paper presents MedReAct'N'MedReFlex, which leverages a suite of four LLM-based medical agents. The MedReAct agent initiates the process by observing, analyzing, and taking action, generating trajectories to guide the search to target a potential error in the clinical notes. Subsequently, the MedEval agent employs five evaluators to assess the targeted error and the proposed correction. In cases where MedReAct's actions prove insufficient, the MedReFlex agent intervenes, engaging in reflective analysis and proposing alternative strategies. Finally, the MedFinalParser agent formats the final output, preserving the original style while ensuring the integrity of the error correction process. One core component of our method is our RAG pipeline based on our ClinicalCorp corpora. Among other well-known sources containing clinical guidelines and information, we preprocess and release the open-source MedWiki dataset for clinical RAG application. Our results demonstrate the central role of our RAG approach with ClinicalCorp leveraged through the MedReAct'N'MedReFlex framework. It achieved the ninth rank on the MEDIQA-CORR 2024 final leaderboard. 
\end{abstract}

\section{Introduction}

\begin{figure*}[!h]
\centering
\includegraphics*[width=0.9\linewidth]{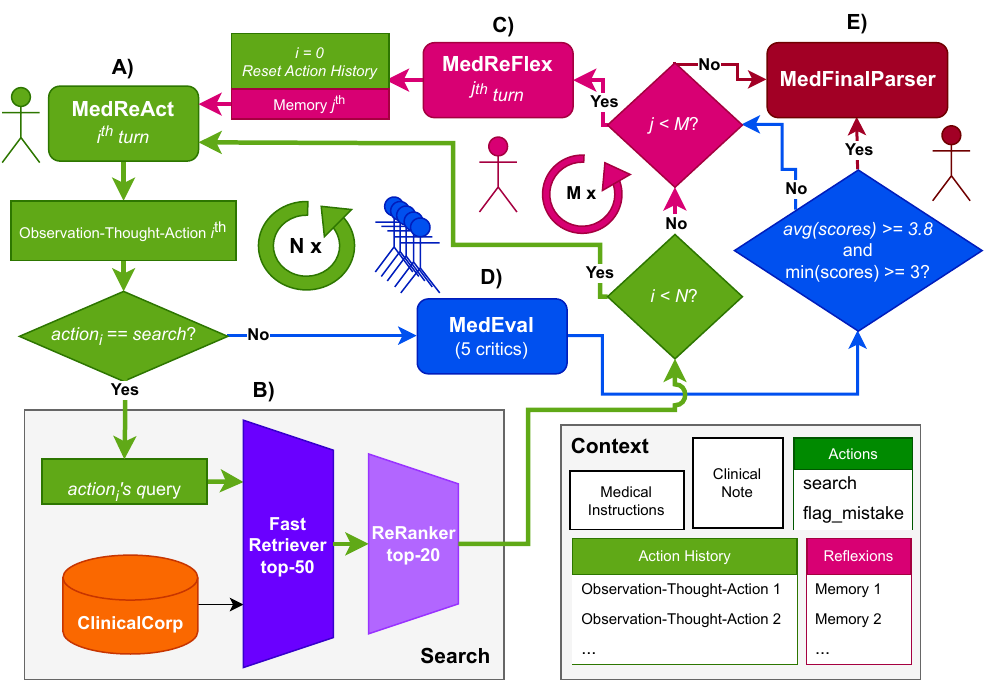}
\caption{Schema of \textit{MedReAct'N'MedReFlex} along the context of the clinical error correction task accessible to all medical agents: \textit{MedReAct}, \textit{MedReFlex}, \textit{MedEval} and \textit{MedFinalParser}. A) The \textit{MedReAct} agent first provides an observation, a thought and an action. B) In the case of a \textit{search} action, it triggers a semantic search over ClinicalCorp using \textit{MedReAct}'s query. Then, the MedReAct agent loops up to $N$ times (green inner loop) or until a \textit{final\_mistake} action is provided. C) After $N$ unsuccessful searches from \textit{MedReAct}, the \textit{MedReFlex} agent reflects on the current situation and suggests a solution (pink outer loop). Then, \textit{MedReAct} might start again. D) Once \textit{MedReAct} selects the \textit{final\_mistake} action, the five \textit{MedEval} agents review the answer and give a score between 1 and 5 (blue line). E) If the average equals or surpasses 3.8 and the minimum above or equal to 3, the \textit{MedFinalParser} agent formats the final answer into a JSON object. If the answer is unsatisfactory, \textit{MedReFlex} is triggered instead. If \textit{MedReFlex} reaches unsuccessfully the $M^{th}$ turns, \textit{MedFinalParser} concludes that there is no error.}
\label{fig:schema}
\end{figure*}

In natural language processing applied to the clinical domain, the accurate detection and correction of medical errors are paramount tasks with profound implications for patient care and safety. This paper introduces the multi-agent framework MedReAct'N'MedReFlex, meticulously engineered to tackle medical error detection and correction, as delineated in the MEDIQA-CORR 2024 competition.

Our framework integrates four distinct types of medical agents: MedReAct, MedReFlex, MedEval, and MedFinalParser, each playing a specialized role in the error identification and rectification process. Drawing inspiration from existing frameworks like ReAct \cite{yao2023react} and Reflexion \cite{shinn2023reflexion}, our framework orchestrates a structured approach to error handling.

Leveraging a Retrieval-Augmented Generation (RAG) framework \cite{lewis2020rag} based on MedRAG \cite{xiong2024benchmarking} and MedCPT \cite{jin2023medcpt}, our approach operates over ClinicalCorp, an extensive corpora curated to encompass crucial clinical guidelines. Additionally, we introduce \textit{MedWiki}, a collection of medical articles from Wikipedia. By integrating these resources, our approach seeks to advance state-of-the-art clinical NLP by offering a comprehensive solution tailored to the intricate nuances of medical error handling. Furthermore, this paper documents the construction and release of \textit{MedWiki}, a substantial repository comprising over 1.3 million article chunks. Additionally, we detail the assembly of the ClinicalCorp, a comprehensive corpus comprising \textit{MedWiki} along with other clinical guideline datasets, such as parts of the MedCorp corpora \cite{xiong2024benchmarking} and parts of the guidelines \cite{chen2023meditron}.

Our main contributions are:

\begin{itemize}
    \item We designed a multi-agent framework named \textit{MedReAct'N'MedReFlex} to solve the medical error detection \& correction task (MEDIQA-CORR 2024) based on four types of medical agents: \textit{MedReAct}, \textit{MedReFlex}, \textit{MedEval} and \textit{MedFinalParser}. We deployed this framework on ClinicalCorp using a retrieval-augmented generation approach.
    \item We released the open-source \textit{MedWiki}\footnote{\url{hf.co/datasets/jpcorb20/medical_wikipedia}}, a version of Wikipedia 2022-12-22 focused solely on medical articles. This RAG-ready dataset contains about 1.3M chunks from more than 150K articles, which represents about 3\% of the original corpus.
    \item We provided the recipe to assemble our large corpora \textit{ClinicalCorp} for RAG applications in the clinical domain, containing more than 2.3M chunks.
    \item We released a RAG-ready version\footnote{\url{hf.co/datasets/jpcorb20/rag_epfl_guidelines}} of the open-source \textit{guidelines} used to pre-train \textit{Meditron} \cite{chen2023meditron}, containing more than 710K chunks across eight open-source datasets.
    \item We released our codebase on GitHub\footnote{\url{github.com/microsoft/iryonlp-mediqa-corr-2024}}.
\end{itemize}

\section{Related Work}

\subsection{Medical Large Language Models}

Since the emergence of ChatGPT by OpenAI in December 2022, the landscape of large language models (LLMs) has witnessed a proliferation of both private and public initiatives, leading to the development of increasingly sophisticated models. OpenAI's journey from the GPT3.5-turbo architecture, as reported by \citet{brownlanguage} and \citet{ouyangtraining}, culminated in the release of GPT-4 and its turbo variant \cite{achiam2023gpt}. Similarly, Google introduced Gemini, available in Nano, Pro, and Ultra configurations \cite{team2023gemini}, alongside its open-source Gemma model \cite{team2024gemma}. Anthropic contributed to this landscape with Claude3, offered in three sizes, ranging from Haiku to Opus. Other notable LLMs include Mistral and Mixtral \cite{jiang2023mistral,jiang2024mixtral}, as well as Llama 2 \cite{touvron2023llama} and Yi \cite{young2024yi}. These general-purpose LLMs, such as GPT-4, have demonstrated solid in-context learning capabilities in the medical field \citet{nori2023can}.

Researchers have developed various open-source LLMs with diverse capabilities in the medical NLP domain. Examples include ClinicalCamel \cite{toma2023clinical}, Med42 \cite{med42}, PMC-Llama \cite{wu2023pmc}, BioMedGPT \cite{zhang2023biomedgpt}, Meditron \cite{chen2023meditron}, Apollo \cite{wang2024apollo}, OpenMedLM \cite{garikipati2024openmedlm}, and BioMistral \cite{labrak2024biomistral}. Google also contributed Med-PaLM 2, a specialized LLM tailored for medical tasks \cite{singhal2023large}.

In this study, we employed OpenAI's GPT-4, specifically version turbo 0125, due to its proven state-of-the-art performances in various domains, its functional capabilities, and its large context window of 128K tokens. These attributes make it an ideal foundation for our approach. For instance, \citet{nori2023can} demonstrated that utilizing in-context learning with GPT-4 --- relying on prompt engineering (i.e. few-shot learning \cite{brownlanguage}, chain-of-thought \cite{wei2022chainfs,kojima2022chainzs}, self-consistency \cite{wang2022selfconsistency} and shuffling multiple choice \cite{ko2020positionbias}) --- achieves state-of-the-art performances on medical question-answering tasks, surpassing specialized models like Med-PaLM 2. We relied on a similar approach as our early baseline for medical error detection and correction, discarding the self-consistency and the shuffling techniques since both do not apply to generative tasks. Nonetheless, we have observed low results from which we hypothesized that this approach using only parametric knowledge is lacking reliable knowledge \cite{mallen2023not,ovadia2023fine,kandpal2023large}, which we addressed by applying agentic methods in a retrieval-augmented generation framework.

\subsection{Agentic Methods}

Researchers have devised several agentic methods to enhance LLMs' responses and reasoning capabilities, such as ReAct \cite{yao2023react}, Reflexion \cite{shinn2023reflexion}, DSPy \cite{khattab2023dspy} and self-discovery \cite{zhou2024self}. Additionally, multi-agent paradigms \cite{wu2023autogen} have found application in the medical domain \cite{tang2023medagents}. Our approach draws inspiration from the Reflexion framework \cite{shinn2023reflexion}, which we adapted into our \textit{MedReFlex} agent. Specifically, we implemented a \textit{MedReAct} agent  --- inspired by the ReAct approach \cite{yao2023react} --- to generate trajectories in our environment. However, this agent realizes its sequence of actions in a different order (i.e., observation, thought, and action), enabling streamlined execution.

Given the reliance of the Reflexion framework on feedback mechanisms, we incorporated an LLM-based metric into our \textit{MedEval} medical agents. Evaluation metrics based on prompting strong LLMs, such as GPT-4 \cite{liu2023g}, have demonstrated a high correlation with human judgment. Similar findings have been reported in the medical NLP literature \cite{xie2023enhancing}. Our evaluation protocol involves prompting five GPT-4 reviewers with task-specific criteria: validity, preciseness, confidence, relevance, and completeness. The average and minimum of their scores are both utilized as success criteria, capturing an unbiased final score and the evaluators' confidence, respectively.

\subsection{Retrieval-Augmented Generation}

Before the advent of LLMs, authors have proposed the retrieval-augmented generation (RAG) framework as a mechanism to incorporate non-parametric memory for knowledge-intensive tasks. This framework, as elucidated by Lewis et al. \cite{lewis2020rag}, leverages both sparse \cite{robertson2009bm25} and dense \cite{reimers2019sentence} retrieval methods. In the medical NLP domain, MedCPT \cite{jin2023medcpt} serves as a prominent retrieval approach, augmented by a reranking stage based on a cross-encoder model. Notably, Xiong et al. \cite{xiong2024benchmarking} conducted a comprehensive study on RAG applications in the medical domain, culminating in developing the MedRAG framework and the MedCorp corpora. Our approach builds upon these foundations, employing the MedCPT retrieval techniques and two corpora from MedCorp.

A pivotal aspect of RAG is its search engine's collection of indexed documents. The \textit{guidelines} corpora, part of the \textit{GAP-replay} corpora, was curated to train Meditron \cite{chen2023meditron}. This corpus comprises web pages describing medical guidelines from reputable healthcare websites like the World Health Organization. The \textit{StatPearls} and \textit{Textbooks} datasets, included in the \textit{MedCorp} corpora used in MedRAG \cite{xiong2024benchmarking}, encompass documents from clinical decision support tools and medical textbooks \cite{jin2021disease}. While \textit{Wikipedia} and \textit{PubMed} datasets within \textit{MedCorp} offer extensive data (i.e. more than 55M documents), we opted for efficiency by focusing on the smaller \textit{PubMed} subset in the \textit{guidelines} corpora and our \textit{MedWiki} corpus.

\section{Methodology}

\subsection{MEDIQA-CORR Task}

The goal of the medical error detection and correction task \cite{mediqa-corr-task} from the clinical note is threefold: detect the presence of an error, locate the sentence containing the error and generate a corrected version of that sentence. The input of the dataset \cite{mediqa-corr-dataset} is a clinical note of several sentences containing a medical description of a patient's condition, test results, diagnosis, treatment and other aspects. There are two parts for the validation and test sets: \textit{MS} from Microsoft and \textit{UW} from the University of Washington. As a primary evaluation metric, the organizers asked to utilize the aggregation score defined by \citet{abacha2023overview} over Rouge-1, BertScore and BLEURT, demonstrating a higher correlation with human judgement.

\begingroup
\setlength{\extrarowheight}{4.5pt}
\begin{table*}[t!]
    \caption{Datasets gathered to construct ClinicalCorp.}
    \label{tab:dataset_summary}
    \centering
    \begin{tabular}{|c|l|l|r|r|}
    \hline
    \small{\textbf{Dataset}} & \small{\textbf{Source}} & \small{\textbf{Status}} & \small{\textbf{\# Documents}} & \small{\textbf{\# Chunks}} \\ \hline
    \multirow{13}{*}{\shortstack[c]{Guidelines\\\cite{chen2023meditron}}} & \small{WikiDoc} & open & 33,058 & 360,070 \\ \cline{2-5}
    & \small{PubMed (guidelines only)} & open & 1,627 & 124,971 \\ \cline{2-5}
    & \small{National Institute for Health and Care Excellence} & open & 1,656 & 87,904 \\ \cline{2-5}
    & \small{Center for Disease Control and Prevention} & open & 621 & 70,968 \\ \cline{2-5}
    & \small{World Health Organization} & open & 223 & 33,917 \\ \cline{2-5}
    & \small{Canadian Medical Association} & open & 431 & 18,757 \\ \cline{2-5}
    & \small{Strategy for Patient-Oriented Research} & open & 217 & 11,955 \\ \cline{2-5}
    & \small{Cancer Care Ontario} & open & 87 & 2,203 \\ \cline{2-5}
    & \small{Drugs.com} & close & 6,711 & 37,255 \\ \cline{2-5}
    & \small{GuidelineCentral} & close & 1,285 & 2,451 \\ \cline{2-5}
    & \small{American Academy of Family Physicians} & close & 60 & 130 \\ \cline{2-5}
    & \small{Infectious Diseases Society of America} & close & 54 & 7,785 \\ \cline{2-5}
    & \small{Canadian Paediatric Society} & close & 43 & 1,123 \\ \hline
    \multirow{2}{*}{\shortstack[c]{MedCorp\\\cite{xiong2024benchmarking}}} & \small{StatPearls} & close & 9,379 & 307,187 \\ \cline{2-5}
     & \small{Textbooks \cite{jin2021disease}} & open & 18 & 125,847 \\ \hline
    \multirow{2}{*}{\shortstack[c]{\textbf{ClinicalCorp}\\(Ours)}} & \small{MedWiki} & open & 150,380 & 1,139,464 \\ \cline{2-5}
    & \small{\textbf{All}} & \textbf{mix} & \textbf{205,850} & \textbf{2,331,987} \\ \hline
    \end{tabular}
\end{table*}
\endgroup

\subsection{ClinicalCorp Corpora}

Our corpus is detailed in Table \ref{tab:dataset_summary}.

\paragraph{guidelines} We aggregated 13 datasets --- which are open-source or closed-source --- from the \textit{guidelines} corpora. We adapted and ran the scrappers from the Meditron GitHub repository to gather the closed-source datasets. Then, we chunked the resulting documents using LangChain's recursive-character text splitter \cite{Chase2022langchain} with a chunk size of 1,000 characters and an overlap of 200 characters, as used for \textit{StatPearls} (see next section).

\paragraph{MedCorp} We gathered two of the four datasets contained in \textit{MedCorp} from MedRAG \cite{xiong2024benchmarking}: \textit{StatPearls} and \textit{Textbooks}. The former was downloaded, cleaned and chunked using \textit{MedRAG} GitHub repository, while the latter was readily available on the \textit{HuggingFaceHub}\footnote{\url{hf.co/datasets/MedRAG/textbooks}}.

\paragraph{MedWiki} We filtered the 2022-12-22 Wikipedia dump\footnote{\url{hf.co/datasets/Cohere/wikipedia-22-12}} pre-processed into chunks by \textit{Cohere} for medical articles only. To select the medical articles, we leveraged an available fine-tuned BerTopic\footnote{\url{hf.co/MaartenGr/BERTopic_Wikipedia}} \cite{grootendorst2022bertopic}, trained on the same Wikipedia dump. We associated its 2,3K topics to the medical domain based on the topics' word representations --- e.g. topic \textit{1850} is related to the medical field, and it corresponds to the word representations: shingles, herpesvirus, chickenpox, herpes, smallpox, zoster, immunity, infectivity, inflammation, and viral. We made these predictions by prompting \textit{GPT3.5-turbo 0613} with a temperature of $1.0$ followed by a majority vote over five predictions. If at least four were positive, we declared the topic medically relevant. In the manual verification of about 50 diverse medical terms on the resulting collection, we observed a near-perfect coverage of Wikipedia's articles related to diseases, treatments, bacteria, or drugs. Only two topics were missing\footnote{Index \textit{509} related to biological taxonomy and \textit{806} related to yeasts.}, corresponding to one single example from the manual test. Given that our goal is to reduce the size of this dataset and use it in an RAG application, we added these topics manually. We obtained a corpus of 150K articles and nearly 1.4M chunks.

\subsection{Semantic Search}

We followed the MedCPT approach \cite{jin2023medcpt} in two stages (see step \textit{B} in Figure \ref{fig:schema}), which is composed of a fast bi-encoder retrieving stage followed by a cross-encoder reranking stage.

We implemented the first stage on a ChromaDB instance, in which we loaded \textit{ClinicalCorp}. This stage aims to find relevant documents while maintaining a good accuracy/latency trade-off. This vector database embeds documents using a fast bi-encoder model \cite{reimers2019sentence}. Then, we provide a query to fetch the closest documents under a given distance, computed with the hierarchical navigable small world approximation (HNSW, by \citet{malkov2018efficient}). We experimented with three bi-encoders from the \textit{HuggingFaceHub}: \textit{sentence-transformers/all-MiniLM-L6-v2} (default), \textit{NeuML/pubmedbert-base-embeddings-matryoshka} and MedCPT original Query/Article encoders. According to our initial experiments, we discarded \textit{all-MiniLM-L6-v2} because we noticed a critical lack of knowledge about medical terminology hindering its accuracy despite a very low latency. NeuML's model and MedCPT's are Bert-based models of 768 hidden dimensions and 12 layers, a slow architecture to generate sentence embeddings. However, NeuML fine-tuned a recent model using the Matryoshka Representation Learning technique \cite{kusupati2022matryoshka}, allowing to truncate dimensions down to 256 dimensions of the 768 embeddings, which significantly accelerated the computations. Our experiments employ this MRL encoder with truncation at 256 dimensions as a trade-off between accuracy and latency.

We implemented the reranking stage following the cross-encoder approach from MedCPT \cite{jin2023medcpt}. Our early experimentation demonstrated the superiority of this model compared to NeuML's MRL bi-encoder with all 768 dimensions as a reranker.

\section{MedReAct'N'MedReFlex Framework}

Unlike previous multi-agent frameworks \cite{wu2023autogen,tang2023medagents}, our approach diverges from a free conversation format to adopt a fixed design schema, as illustrated in Figure \ref{fig:schema}. Within this structured framework, each medical agent intervenes at a specific step, facilitating a systematic and coordinated approach to address the error detection and correction task. Central to our methodology are four distinct medical agents: MedReAct, MedReFlex, MedEval, and MedFinalParser.

\subsection{MedReAct Agent}
The MedReAct agent (see step \textit{A} in Figure \ref{fig:schema}), inspired by the ReAct framework \cite{yao2023react}, operates cyclically, beginning with an observation of the current context, followed by a thoughtful analysis, and concluding with an action (\textit{search} or \textit{final\_mistake}). This agent generates a trajectory of up to $N$ steps if the action is always a \textit{search} with different queries. 

We also experimented with adding two other actions (\textit{get\_doc\_by\_id} and \textit{next\_results\_from\_query}), but MedReAct systematically ignored them.

\subsection{MedEval Agent}
Upon MedReAct's selection of the \textit{final\_mistake} action, the MedEval agents (see step \textit{D} in Figure \ref{fig:schema}), akin to the GPT-Eval approach \cite{liu2023g}, evaluate the proposed solution. Five GPT-4-based evaluators assess the answer based on criteria such as validity, preciseness, confidence, relevance, and completeness. The ensemble of evaluators ensures comprehensive and unbiased feedback, contributing to robust error detection and correction. We leverage the average final score as well as the minimum review score. We added this condition on the minimum score to capture the confidence of the evaluation. If one reviewer gave a much lower score than the others, we experimentally observed that it was often a signal of lower confidence in the final answer.

\subsection{MedReFlex Agent}
In scenarios where MedReAct's actions fail to yield satisfactory outcomes, the MedReFlex agent (see step \textit{C} in Figure \ref{fig:schema}), drawing from the Reflexion framework \cite{shinn2023reflexion}, intervenes. This agent engages in reflective analysis to reassess the situation. By considering contextual cues, past interactions and all five reviews, MedReFlex proposes alternative strategies to address the identified challenges. This iterative process allows for adaptive decision-making and fosters resilience in error detection and correction tasks.

\subsection{MedFinalParser Agent}
Suppose the average score provided by the MedEval agents exceeds or equals 4, and the minimum score surpasses or equals 3. In that case, the MedFinalParser agent (see step \textit{D} in Figure \ref{fig:schema}) proceeds to format the final answer into a JSON object. This agent also ensures the conservation of the original style of the clinical note, which the MedReAct agent tends to disrupt by copying the writing style of the search documents. Conversely, if the answer falls short of the predetermined thresholds, MedReFlex is triggered for further refinement. If MedReFlex's interventions prove ineffective after the $M^{th}$ turn, the MedFinalParser agent concludes that no errors exist, ensuring the integrity of the error correction process.

\section{Results}

\subsection{Results for the Competition}

MedReAct'N'MedReFlex achieved the $9^{th}$ rank during the MEDIQA-CORR 2024 official testing period, corresponding to an aggregation score of $0.581$. Nonetheless, we thoroughly optimize our method in the following sections. To complete these experiments in a reasonable amount of time, we randomly sample 50 examples from the MS validation set.

\subsection{Agentic Method Comparison}

In Table \ref{tab:agentic_results}, we compared the MedReAct agent only against using our proposed method MedReAct'N'MedReFlex. Our approach achieves more than a few absolute percent across metrics. We also experimented with a baseline inspired from \citet{nori2023can} (i.e. "MedPrompt") with in-context learning prompting alone, but the results were drastically lower.

\begin{table}[h!]
    \centering
    \begin{tabular}{|c|c|c|}
        \hline
        \small{Metric} & \textbf{\small{MedReAct}} & \textbf{\small{MedReAct'N'MedReFlex }}\\
        \hline
        \small{ROUGE-1} & 0.504 & 0.568\\
        \hline
        \small{BERTScore} & 0.580 & 0.642\\
        \hline
        \small{BLEURT} & 0.531 & 0.588\\
        \hline
        \small{Aggregate} & 0.539 & 0.599\\
        \hline
    \end{tabular}
    \caption{Comparison between MedReAct agent only with up to 10 turns against MedReAct'N'MedReFlex with 4 turns for MedReAct and 5 turns for MedReFlex, leveraging the optimal search configuration (retrieval top-k at 50 and reranking top-k at 20).}
    \label{tab:agentic_results}
\end{table}

\subsection{Semantic Search Optimization}

After the end of the MEDIQA-CORR 2024 shared task, we carried out a thorough analysis of our semantic search engine. The main parameters to tune are retrieval top-k, reranking top-k and the source included in ClinicalCorp.

\subsubsection{Retrieval Top-K}

\begin{figure}[!h]
\centering
\includegraphics*[width=\linewidth]{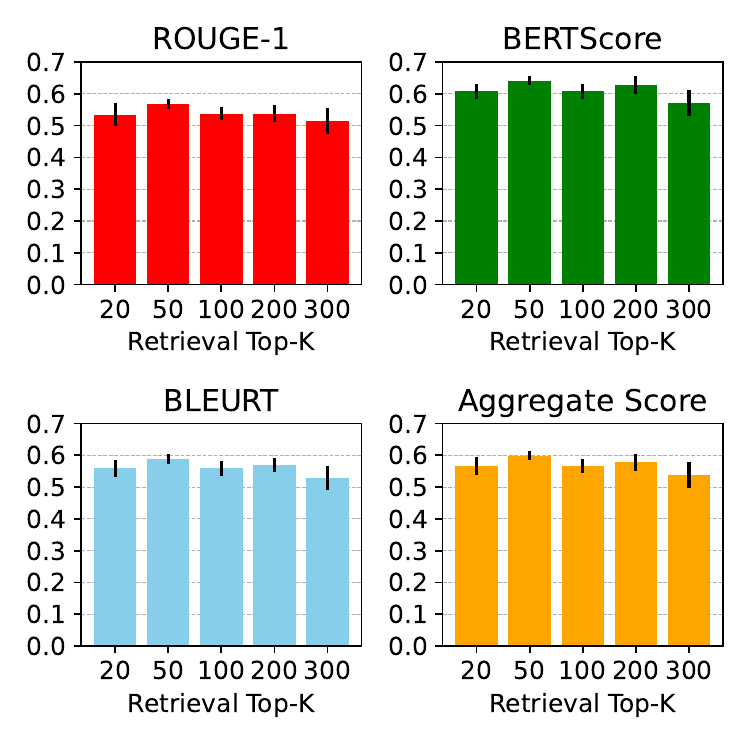}
\caption{Performances across many retrieval top-k values with a reranking top-k set at 20 over 3 runs.}
\label{fig:retrieval_topk}
\end{figure}

In Figure \ref{fig:retrieval_topk}, we illustrate the performances across many retrieval top-k values employing a fixed reranking top-k of 20. For the official ranking of the MEDIQA-CORR 2024, we set this value to 300. However, we observe here that this setting is sub-optimal. A retrieval top-k of 50 improves the final performances by a few absolute percent. We interpret this observation as indicative of a misalignment between our task and the fine-tuning of the MedCPT reranker. The more documents we provide to the reranking model (e.g. 200 or 300), the more low-relevance documents are put in the top 20 by the reranker output.

\begin{figure}[!h]
\centering
\includegraphics*[width=\linewidth]{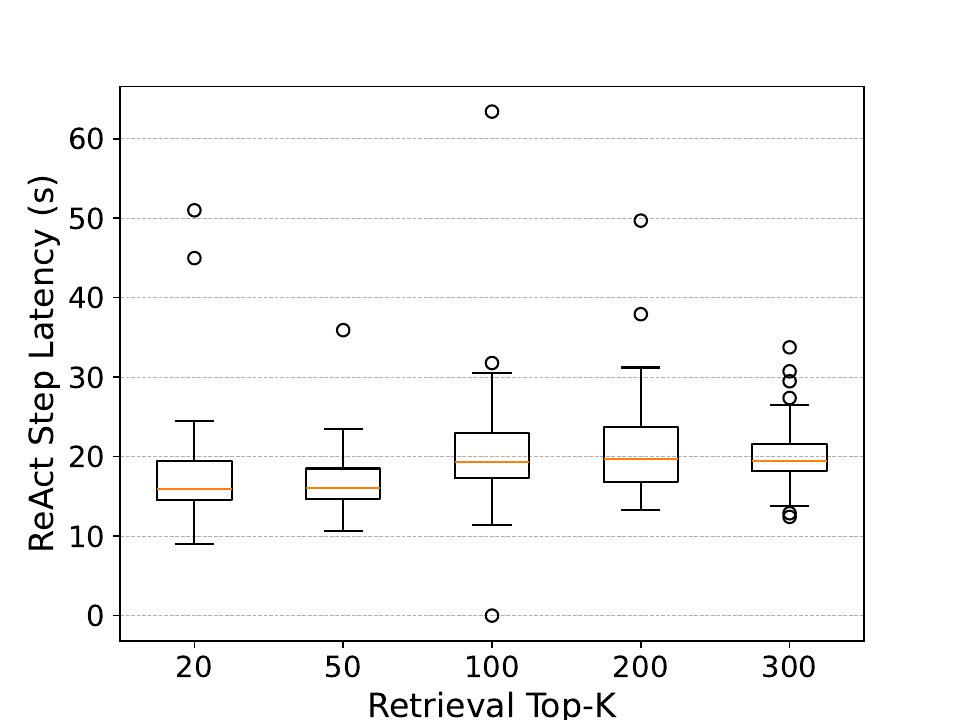}
\caption{ReAct step average latency per retrieval top-k with a reranking top-k set at 20.}
\label{fig:retrieval_topk_latency}
\end{figure}

Nonetheless, a reranking without surplus documents --- i.e. retrieval top-k of 20 with a reranking top-k of 20 --- remains sub-optimal, mainly in contrast to using 50 documents. In Figure \ref{fig:retrieval_topk_latency}, we provide the associated average latency for one react step in seconds. We notice that the latency seems to scale with the order of magnitude of the retrieval top-k, with a value of 20 and 50 having 17 seconds on average, while 100, 200 and 300 are around 20 seconds. We expected that the reranking of 300 examples against 100, for instance, would lead to noticeable latency, but it is negligible in contrast to the retrieval from ChromaDB over our 2.3M chunks.

\begin{figure}[!h]
\centering
\includegraphics*[width=\linewidth]{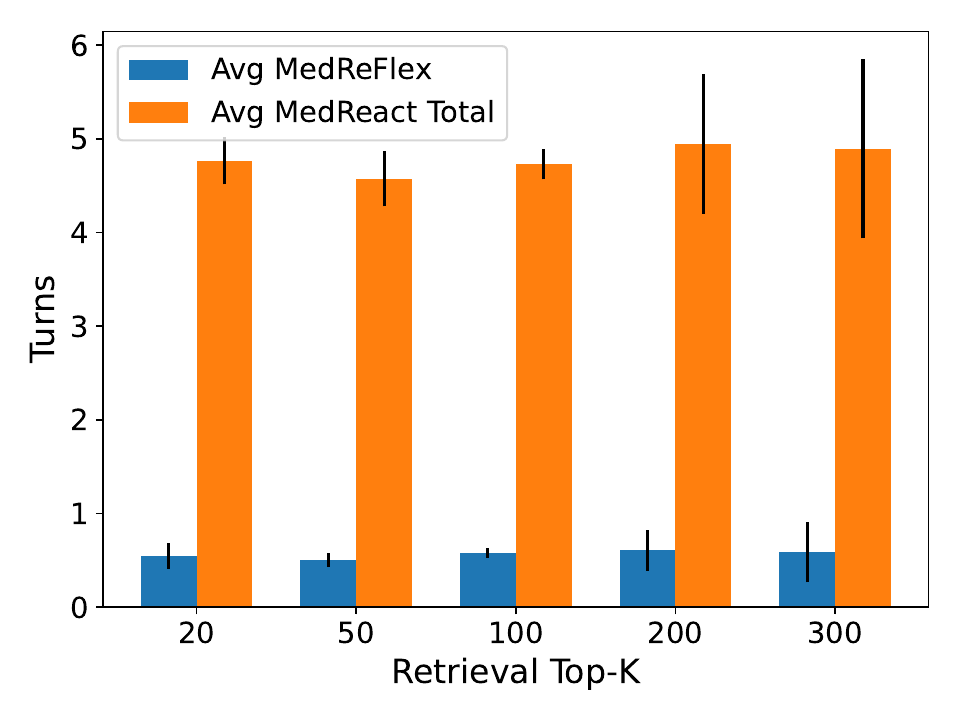}
\caption{Average turns of MedReAct and MedReFlex according to various retrieval top-k with a reranking top-k set at 20.}
\label{fig:retrieval_topk_turns}
\end{figure}

We also show the average amount of MedReFlex turns and the average of the sum of all MedReAct turns in Figure \ref{fig:retrieval_topk_turns}. Overall, the trends are similar, with 4.8 total ReAct turns on average, but there is a slight increase in the average and variance for top-k values of 200 and 300. Therefore, these settings are underperforming and slower regarding latency and the number of turns needed to reach an answer.

Overall, the retrieval top-k of 50 leads to higher performances across all metrics and reduced latency and number of turns required by our algorithm.

\subsubsection{Reranker Top-K}

\begin{figure}[!h]
\centering
\includegraphics*[width=\linewidth]{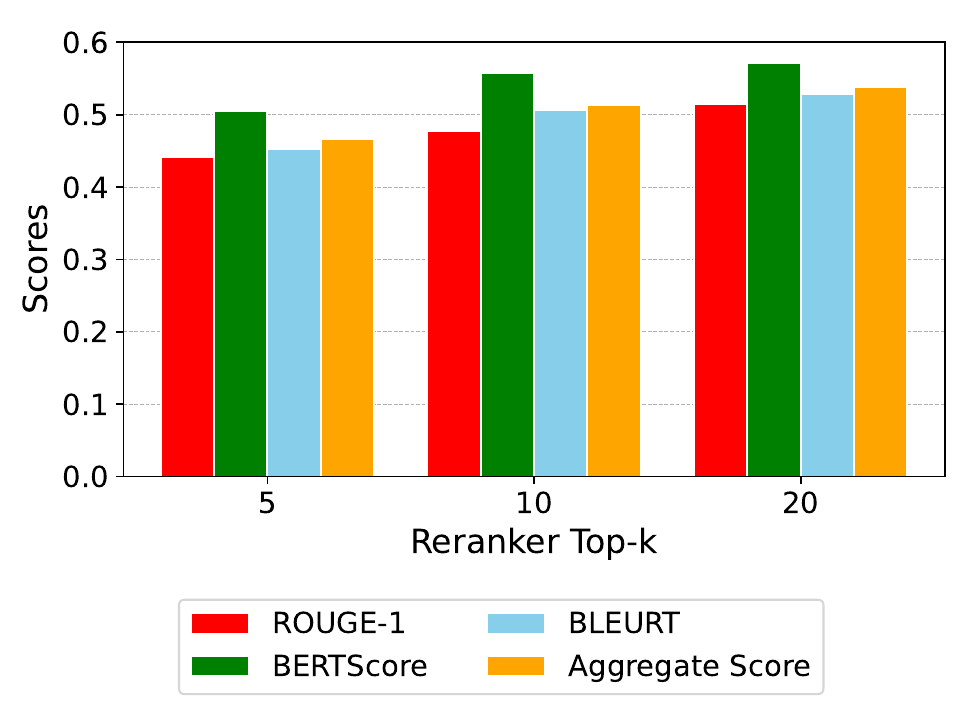}
\caption{Reranker top-K with a retrieval top-k set at 300.}
\label{fig:reranker_topk}
\end{figure}

In Figure \ref{fig:reranker_topk}, we fix the retrieval top-k at 300 and compute the performances across three reranking top-k values: 5, 10 and 20. Since the context window of the LLM limits us, we constraint the maximum of $K$ to 20, given that these $K$ documents are injected in the prompt up to $N$ times for each MedReAct step. According to Figure \ref{fig:reranker_topk}, we observe that the more documents we provide in the prompt, the more we increase the performances --- the aggregate score gains close to 10\% absolute when augmenting from 5 to 20 documents.

\subsubsection{Sources in ClinicalCorp}

\begin{figure}[!h]
\centering
\includegraphics*[width=\linewidth]{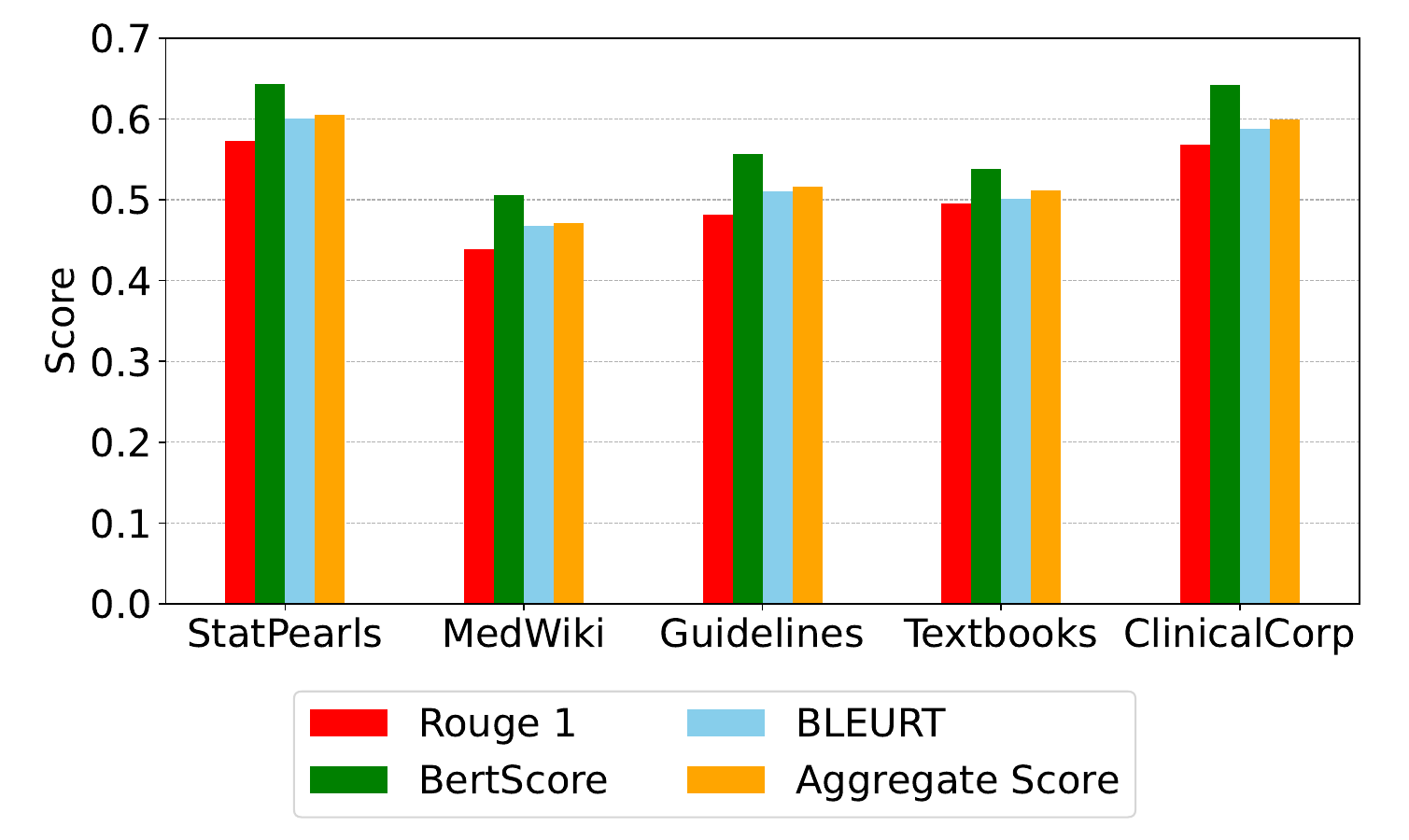}
\caption{Performances per source in ClinicalCorp with the retrieval top-k set at 50 and the reranking top-k set at 20.}
\label{fig:scoure_results}
\end{figure}

We measure the impact of each source in ClinicalCorp in Figure \ref{fig:scoure_results}. First, we observe that \textit{MedWiki} is the lowest-performing source of documents with an aggregation score of nearly 0.47. \textit{guidelines} and \textit{Textbooks} provide a similar accuracy at about 0.51 in aggregate score. Finally, \textit{StatPearls} leads to the highest score close to the full ClinicalCorp. Given our small validation set of 50 examples, we consider it a better practice to keep all ClinicalCorp for our task since more edge cases might appear at test time.

\begin{figure}[!h]
\centering
\includegraphics*[width=\linewidth]{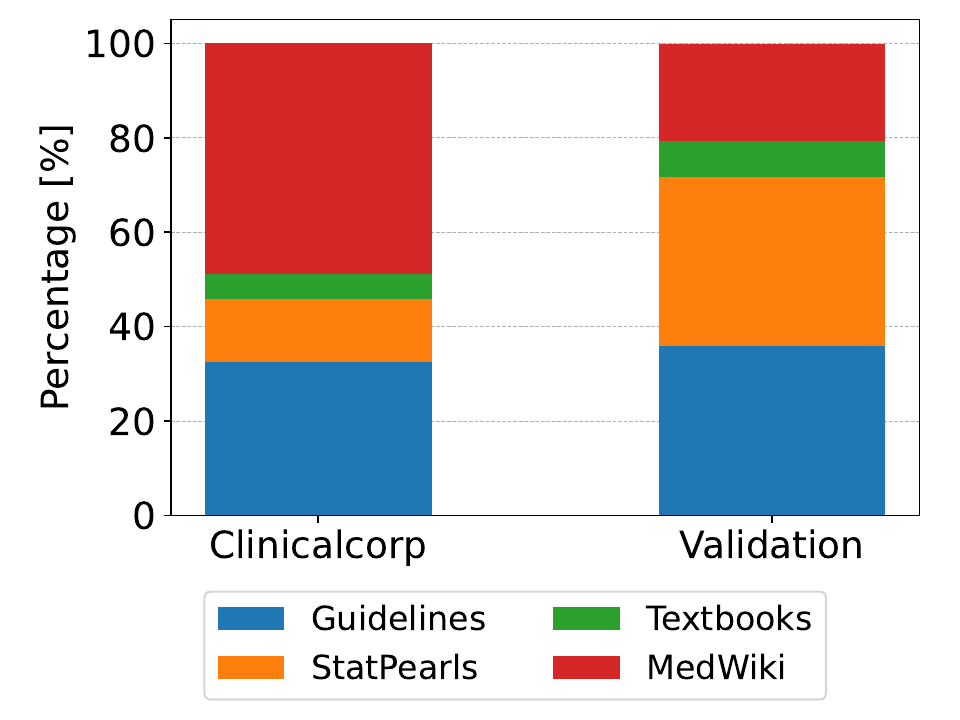}
\caption{Distribution of sources' chunks in ClinicalCorp against appearances of these chunks' sources in one run of MedReAct'N'MedReFlex.}
\label{fig:scoure_dist}
\end{figure}

We show in Figure \ref{fig:scoure_dist} the distributions of sources from ClinicalCorp in general in comparison to the distributions of sources' chunks used by one run of MedReAct'N'MedReFlex. We observe, in general, a much larger utilization of the \textit{StatPearls}' chunks in contrast to the \textit{MedWiki}'s chunks, while we remark similar distributions for the other two datasets. These results align with the previous analysis demonstrating a higher performance from using only \textit{StatPearls}.

\subsection{MedEval Aggregation Thresholds}

\begin{figure}[!h]
\centering
\includegraphics*[width=\linewidth]{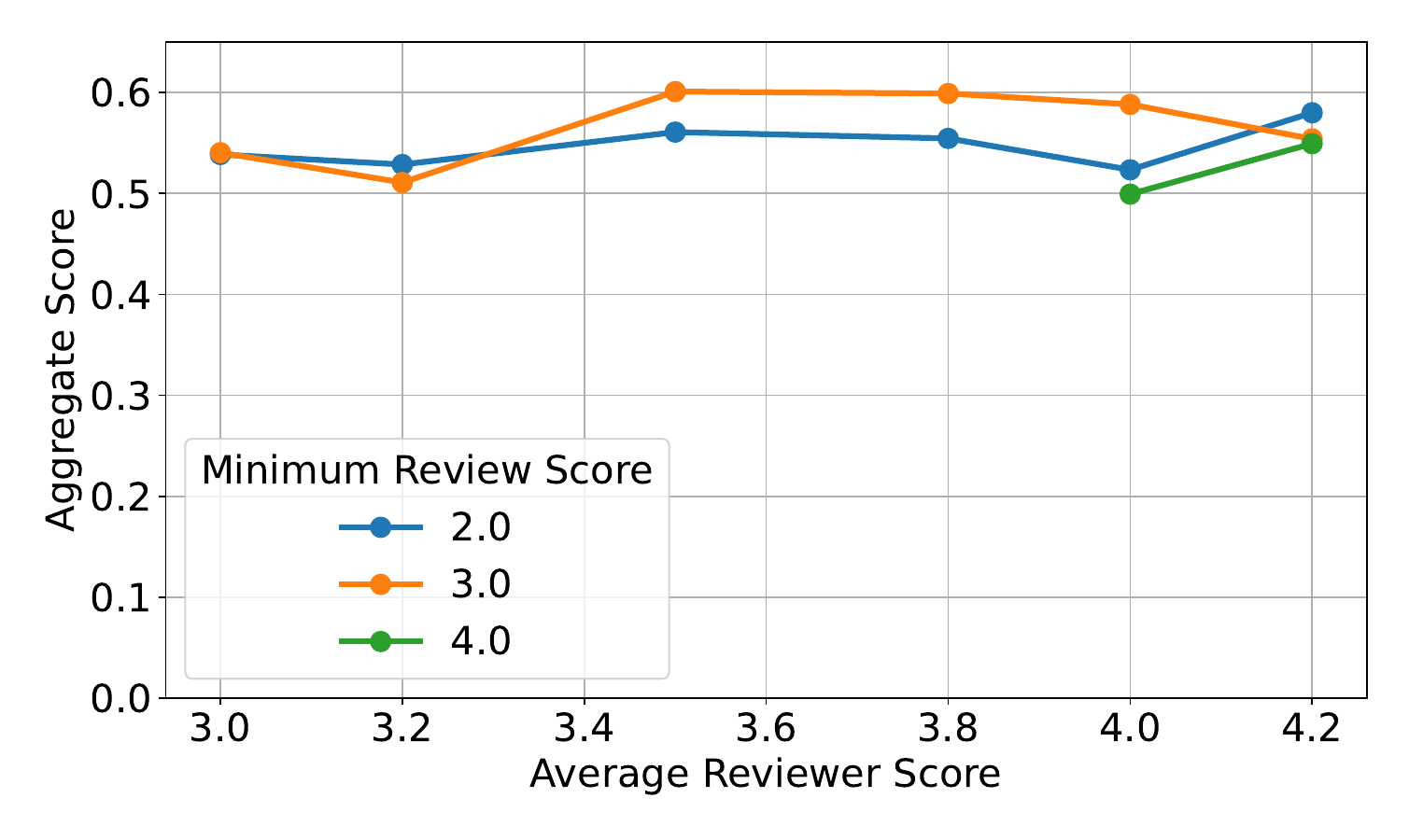}
\caption{Aggregate scores across different MedEval's average and minimum thresholds with the retrieval top-k set at 50 and the re-ranker top-k set at 20. We omitted average thresholds that are strictly lower for consistency for a given minimum threshold.}
\label{fig:review_analysis}
\end{figure}

In Figure \ref{fig:review_analysis}, we show the impact of applying different thresholds to the average and minimum review scores on the performance. For the minimum score criterion, we choose the integer values of $2.0$, $3.0$ and $4.0$. We select values for the average score criterion: $3.0$, $3.2$, $3.5$, $3.8$, $4.0$ and $4.2$. We do not compute the performances for combinations where the minimum threshold is higher than the average threshold for mathematical consistency. We observe an optimal setting for a minimum evaluation score of $3.0$ with a range of average evaluation scores in $\left[3.5, 3.8\right]$.

\section{Conclusion}

In this paper, we introduced MedReAct'N'MedReFlex, a multi-agent framework developed for the MEDIQA-CORR 2024 competition aimed at medical error detection and correction in clinical notes. The framework incorporates four specialized medical agents: MedReAct, MedReFlex, MedEval, and MedFinalParser, leveraging the RAG framework and our ClinicalCorp. We detail the construction of our ClinicalCorp, including diverse clinical datasets such as \textit{guidelines}, \textit{Textbooks}, and \textit{StatPearls}. Additionally, we released MedWiki, a corpus comprising Wikipedia medical articles. Our framework achieved the ninth rank in the competition with an aggregation score of 0.581. Through optimization experiments, we identified sub-optimal settings at the time, demonstrating substantial performance improvements with a retrieval top-k of 50, a reranking top-k of 20, an average review threshold of 3.8, and a minimum review threshold of 3. As future work, we envision refining the chunking strategy on the ClinicalCorp, applying further prompt engineering of the medical agents, and conducting a deeper analysis of the interactions between the MedReAct'N'MedReFlex's agents.
\section*{Acknowledgments}

We would like to thank Jean-Michel Attendu, François Beaulieu, and Paul Vozila from Microsoft Health \& Life Sciences team for their support, as well as the ClinicalNLP workshop's organizers, reviewers and other participants. 

\bibliography{custom}

\end{document}